\documentclass[11pt]{article}

\usepackage[final]{acl}

\usepackage{times}
\usepackage{latexsym}

\usepackage[T1]{fontenc}
\usepackage[utf8]{inputenc}

\usepackage{microtype}

\usepackage{inconsolata}

\usepackage{graphicx}
\usepackage{amsmath}
\usepackage{amssymb}
\usepackage{amsthm}
\usepackage{booktabs}
\usepackage[english]{babel}
\usepackage{makecell}
\usepackage{algorithm}
\usepackage{algorithmic}
\usepackage{multirow} 
\usepackage{subcaption}
\usepackage{pifont}
\usepackage[table]{xcolor}
\newcommand{\cco}[1]{\cellcolor{orange!25}#1}  % 第二好(橙色)
\newcommand{\ccr}[1]{\cellcolor{red!25}#1}     % 最好(红色)
\title{Expert Pyramid Tuning: Efficient Parameter Fine-Tuning for Expertise-Driven Task Allocation}

\author{
 \textbf{Jia-Chen Zhang\footnotemark[2]},   % [2] 自动显示为 †
 \textbf{Zhen-Wei Yan\footnotemark[2]},
 \textbf{Yu-Jie Xiong\footnotemark[1]},     % [1] 自动显示为 *
 \textbf{Chun-Ming Xia\footnotemark[1]}
\\
 School of Electronic and Electrical Engineering, Shanghai University of Engineering Science \\
 333 Longteng Road, Shanghai, China
\\
 \small{
   \textbf{Correspondence:} 
   \href{mailto:xiong@sues.edu.cn}{xiong@sues.edu.cn}, 
   \href{mailto:cmxia@sues.edu.cn}{cmxia@sues.edu.cn}
 }
}
\begin{document}
\maketitle
\footnotetext[1]{Corresponding author.}
\footnotetext[2]{These authors contributed equally to this work.}
\begin{abstract}
Parameter-Efficient Fine-Tuning (PEFT) has become a dominant paradigm for deploying LLMs in multi-task scenarios due to its extreme parameter efficiency. While Mixture-of-Experts (MoE) based LoRA variants have achieved promising results by dynamically routing tokens to different low-rank experts, they largely overlook the hierarchical nature of task complexity. Existing methods typically employ experts with uniform architectures, limiting their ability to capture diverse feature granularities required by distinct tasks—where some tasks demand high-level semantic abstraction while others require fine-grained syntactic manipulation. To bridge this gap, we propose Expert Pyramid Tuning (EPT), a novel architecture that integrates the multi-scale feature pyramid concept from computer vision into the realm of PEFT. Unlike standard LoRA, EPT decomposes task adaptation into two stages: (1) A shared meta-knowledge Subspace that encodes universal linguistic patterns in low dimensions; (2) A Pyramid Projection Mechanism that utilizes learnable up-projection operators to reconstruct high-dimensional features at varying scales. A task-aware router then dynamically selects the optimal combination of these multi-scale features. Extensive experiments across multiple multi-task benchmarks demonstrate that EPT significantly outperforms SOTA MoE-LoRA variants. Crucially, thanks to the re-parameterization capability of our design, EPT achieves this performance improvement while simultaneously reducing the number of training parameters. Our code is publicly available at: \url{https://anonymous.4open.science/r/EPT-B0E4}.
\end{abstract}

\begin{table}[htbp]
  \centering
  \renewcommand\arraystretch{1.0}
  \begin{tabular}{l|cccc}
    \toprule
    Rank & MRPC & RTE & SST-2 & CoLA \\
    \midrule
    1  & \textbf{89.7} & 77.6 & 94.4 & 60.9 \\
    2  & 89.2 & 78.7 & 94.6 & 60.0 \\
    4  & 88.7 & \textbf{80.5} & \textbf{94.8} & 61.9 \\
    8  & 89.2 & 77.6 & 94.5 & \textbf{63.3} \\
    16 & 89.2 & 80.1 & 94.4 & 62.3 \\
    32 & 89.5 & 79.1 & 94.5 & 60.5 \\
    \bottomrule
  \end{tabular}
  \caption{LoRA-based Fine-tuning Performance of T5-base with varying ranks on different tasks.}
  \label{tab:1}
\end{table}
\section{Introduction}
Large Language Models (LLMs) have demonstrated remarkable generalization capabilities across a wide spectrum of natural language processing tasks \cite{openai2024gpt4technicalreport, touvron2023llama2openfoundation}. However, adapting these general-purpose models to specific downstream scenarios remains a significant challenge, particularly in multi-task settings. Full fine-tuning is often computationally prohibitive and storage-intensive due to the immense scale of parameters \cite{full-finetuning}. Consequently, Parameter-Efficient Fine-Tuning (PEFT) has emerged as a dominant paradigm \cite{peft, p-tuning, llama-adapter}. Among PEFT techniques, LoRA \cite{hu2022lora} has gained widespread adoption by freezing the pre-trained weights and optimizing low-rank decomposition matrices, thereby striking a favorable balance between adaptation performance and resource efficiency.

Despite its success, standard LoRA struggles to handle the conflicting gradients often inherent in multi-task learning, leading to significant performance degradation known as "negative transfer." To mitigate this, recent studies have integrated the Mixture-of-Experts (MoE) architecture into LoRA \cite{liu2024moe, luo2024moeloracontrastivelearningguided,dou-etal-2024-loramoe}, utilizing gating mechanisms to dynamically route tokens to different low-rank experts. While promising, these methods typically overlook a fundamental characteristic of language processing: the hierarchical nature of task complexity. Existing MoE-LoRA variants predominantly employ experts with a uniform architecture (i.e., identical rank and capacity) \cite{gao-etal-2025-mola, lin-etal-2025-teamlora}. This "one-size-fits-all" design is suboptimal, as different tasks require feature adaptation at varying granularities, as demonstrated by the results in Table \ref{tab:1}. For instance, simple tasks may only require high-level semantic abstraction, whereas complex reasoning or syntactic parsing often demands fine-grained manipulation of representations. Forcing uniformly structured experts to handle such diverse complexities restricts the model's expressiveness and parameter efficiency.

To address this limitation, we draw inspiration from the multi-scale feature hierarchies widely successful in Computer Vision, such as Feature Pyramid Networks (FPN) \cite{lin2017featurepyramidnetworksobject}. In visual processing, recognizing objects at different scales requires capturing features at varying resolutions. We posit that a similar principle applies to parameter-efficient tuning: effective multi-task adaptation requires a "Parameter Pyramid" capable of reconstructing features at multiple levels of granularity. Instead of learning independent, redundant matrices for every expert, we argue that task adaptation should be decomposed into a universal linguistic basis and a task-specific scale projection. Specifically, we first optimize a low-dimensional LoRA incremental matrix. Treating deconvolution layers as experts, we train multiple deconvolutional modules with varying dimensions to project this low-dimensional linguistic basis onto different scales. To align the projected matrices with the dimensions of the frozen pre-trained parameters, EPT incorporates a novel Adaptive LoRA Pruner. This design enables experts across different tasks to share communal knowledge while preserving unique, task-specific information. Furthermore, to fully exploit the discrepancies and correlations among tasks for accurate expert selection, we develop a contrastive learning-based Task Embedding Module. This module assigns a dedicated embedding to each task and employs contrastive optimization to ensure the quality and discriminative power of the learned representations.
We conducted experiments on a wide range of benchmarks to verify the effectiveness of EPT. Our main contributions are as follows:
\begin{itemize}
\item[$\bullet$] We propose Expert Pyramid Tuning (EPT), a novel parameter-efficient framework that introduces the concept of multi-scale feature hierarchies to LoRA-based MoE. By constructing a expert pyramid, EPT dynamically allocates representational capacity based on task complexity, effectively mitigating negative transfer while maintaining high parameter efficiency.
\item[$\bullet$] To ensure compatibility with frozen pre-trained weights, we introduce an Adaptive LoRA Pruner, which aligns the projected multi-scale features with the model's intrinsic dimensions, allowing for flexible and granular feature adaptation.
\item[$\bullet$] We develop a contrastive learning-based task embedding module to optimize expert routing. This mechanism ensures precise expert selection, enabling the model to better distinguish between conflicting tasks and share knowledge across correlated ones.
\item[$\bullet$] Extensive experiments on diverse benchmarks demonstrate that EPT significantly outperforms SOTA PEFT and MoE-LoRA baselines. Our results confirm that EPT not only achieves superior performance in multi-task settings but also exhibits better parameter efficiency and robustness.
\end{itemize}

\begin{figure*}
\centerline{\includegraphics[width=\textwidth]{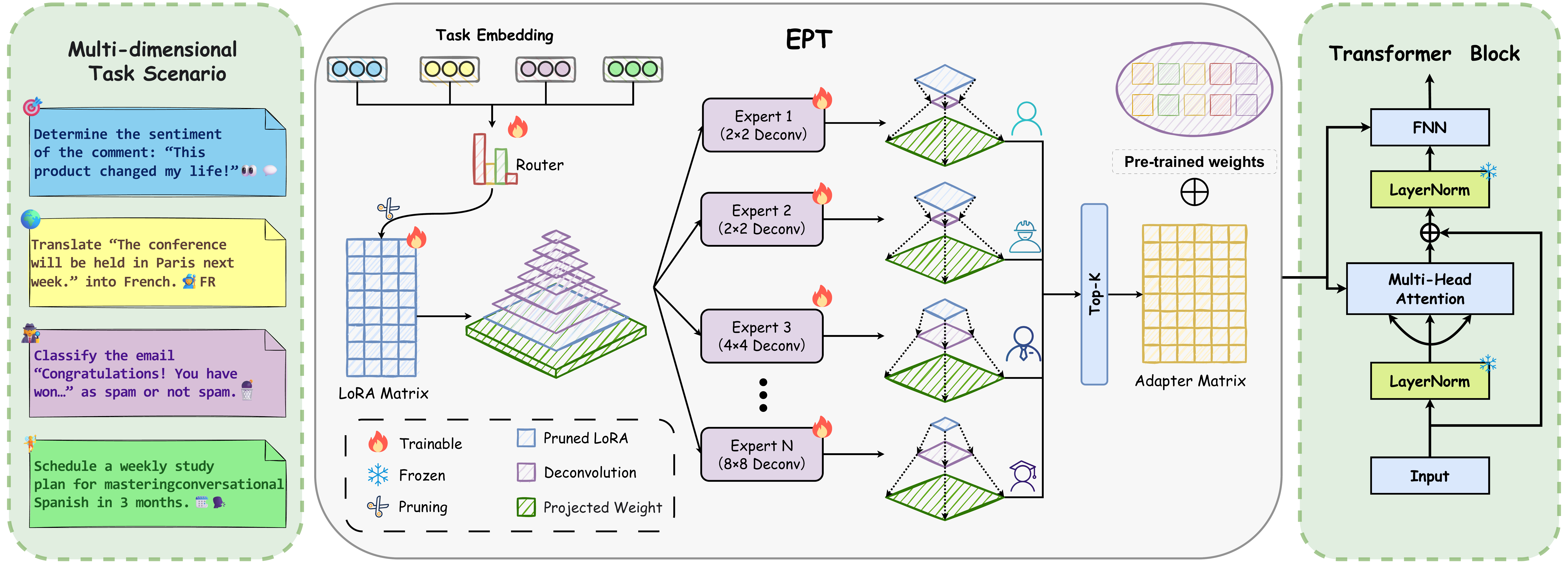}}
  \caption{The overall framework of EPT. The overall architecture of EPT resembles a parameter pyramid, consisting of a shared meta-knowledge subspace and multiple deconvolutional projection layers of varying dimensions.}
  \label{photomain}
\end{figure*}

\section{Related Work}
\subsection{Parameter-Efficient Fine-Tuning}
PEFT has become crucial for adapting LLMs with minimal computational cost \cite{he2022towards}. Among them, LoRA \cite{hu2022lora} has taken a dominant position in this field due to its efficient inference performance. Its core principle lies in indirectly updating the model weights by introducing trainable low-rank matrices, which can then be merged back into the original network during inference. This approach enables fine-tuning of the model with almost no additional latency. Recent improvements to LoRA generally fall into two categories: stability and scaling optimizations, such as DoRA \cite{pmlr-v235-liu24bn} and rsLoRA \cite{kalajdzievski2023rankstabilizationscalingfactor}; and resource-efficiency improvements, such as the quantization-based QLoRA \cite{NEURIPS2023_1feb8787} and subspace deconvolution in DCFT \cite{zhang-etal-2025-parameter}. Attempts have also been made to introduce dynamic rank allocation. For instance, DyLoRA \cite{valipour-etal-2023-dylora} and AdaLoRA \cite{zhang2023adaptive} explore dynamic rank training and budget allocation, respectively. However, these methods are predominantly designed for single-task adaptation and often incur substantial computational overheads due to complex implementation. They fail to address the core challenge of multi-task scenarios, where task complexity varies significantly. Consequently, it is imperative to develop a unified PEFT framework capable of dynamic rank adaptation across multiple tasks, all while maintaining strict training efficiency.

\subsection{Mixture of LoRA}
Recent advancements integrate the MoE paradigm \cite{6797059} into PEFT by treating LoRA modules as experts. These architectures leverage token-level routing to scale capacity while reducing training and inference costs. Foundational mechanisms, such as sparsely-gated top-k routing \cite{shazeer2017, JMLR:v23:21-0998} and auxiliary load-balancing losses \cite{lepikhin2021gshard}, are adapted to optimize expert utilization; others employ Expert-Choice routing \cite{NEURIPS2022_2f00ecd7} to invert the selection process. In this domain, MoELoRA \cite{liu2024moe, luo2024moeloracontrastivelearningguided} and AlphaLoRA \cite{qing-etal-2024-alphalora} focus on contrastive guidance and layer-wise quality allocation. To mitigate redundancy, MoLA \cite{gao-etal-2025-mola} and HydraLoRA \cite{NEURIPS2024_123fd8a5} optimize expert allocation and fusion. Furthermore, MixLoRA \cite{li2024mixloraenhancinglargelanguage} targets high-throughput inference, while MoRE \cite{zhang2025moremixturelowrankexperts} and SMoRA \cite{zhao2025rankexpertsinglerankedmixture} advocate for fine-grained, rank-level expert sharing. Furthermore, several MoE-based fine-tuning models—such as Med-MoE-LoRA, FSMSE and HMVLM \cite{yang2026specializedgeneralistsmultitaskmoelora, cao2026parameterefficientmoelorafewshot, hu2025hmvlmhumanmotionvisionlanuagemodel}—have been developed to address domain-specific multi-task learning. However, these Multi-LoRA schemes remain constrained by adapter duplication and selection overhead, leading to parameter inflation and suboptimal latency. We introduce EPT, a lightweight design leveraging a a shared meta-knowledge subspace and a pyramid projection mechanism to achieve a superior trade-off between accuracy and efficiency.

\section{Method}

\subsection{Expert Pyramid Tuning}
The overall pipeline of EPT is illustrated in Figure \ref{photomain}. To overcome the parameter redundancy inherent in independent expert learning, we propose a decomposition strategy that separates shared meta-knowledge from task-specific structural adaptations. We conceptualize the vast parameter space of large language models as a manifold where diverse tasks share a common low-dimensional latent structure. Accordingly, we explicitly construct a shared meta-knowledge Subspace, denoted as $\mathbf{Z}_{meta} \in \mathbb{R}^{h \times w}$, where $h, w \ll d_{model}$. This subspace is designed to encode universal linguistic patterns in low dimensions and is initialized as a learnable parameter shared across all tasks and experts. Specifically, $\mathbf{Z}_{meta}$ is defined as the product of two low-rank matrices:

\begin{equation}
\mathbf{Z}_{meta} = \mathbf{B} \cdot \mathbf{A},
\end{equation}
where $\mathbf{A}$ and $\mathbf{B}$ are learnable low-rank projection matrices. While conventional LoRA-style methods typically zero-initialize one of the matrices, we employ a random Gaussian distribution for both $\mathbf{A}$ and $\mathbf{B}$. This ensures that the meta-knowledge seed $\mathbf{Z}_{meta}$ captures a rich, non-degenerate latent representation from the onset of training. Unlike traditional methods that directly learn full-rank or low-rank matrices for each expert, EPT constructs a parameter pyramid by projecting this subspace into varying granularities. We define a set of $N$ deconvolutional experts, where the $i$-th expert is parameterized by a unique kernel tensor $\mathcal{K}_i$. Crucially, to emulate the multi-scale hierarchy of visual processing, each kernel $\mathcal{K}_i$ is assigned a distinct spatial configuration (i.e., kernel size $s_i$). This design ensures that each expert captures feature dependencies at a specific scale. Let $\operatorname{Deconv}(\cdot)$ denote the transposed convolution operator. The projection process for the $i$-th expert is defined as:
\begin{equation}
\mathbf{W}_i = \operatorname{Deconv}(\mathbf{Z}_{meta}; \mathcal{K}_i).
\end{equation}
To ensure that the initial output of these experts does not perturb the pre-trained weights, we initialize each kernel $\mathcal{K}_i$ to zero. $\mathbf{Z}_{meta}$ serves as the seed of meta-knowledge, reconstructed into high-dimensional features. To ensure efficiency and structural consistency, we set the deconvolution stride to $s_i$ and employ an Adaptive LoRA Pruner to maintain a uniform output dimensionality across experts. Consequently, the resulting set of matrices $\{\mathbf{W}_1, \dots, \mathbf{W}_N\}$ forms a parameter pyramid. To adaptively combine these multi-scale features, EPT employs a Top-$k$ routing mechanism. Given an input $x$, the gating score $G(x)_i$ for the $i$-th expert is defined as:
\begin{equation}
G(x)_i = \frac{\exp(r_i / \tau)}{\sum_{j \in \mathcal{P}} \exp(r_j / \tau)}, \quad r = \mathbf{W}_r \cdot x,
\end{equation}
where $\mathcal{P}$ is the set of indices corresponding to the top-$k$ values in $r$, and $\mathbf{W}_r$ is the router weight, and $\tau$ is the temperature parameter that controls the smoothness of the gating distribution. The final weight for the EPT layer is then computed as a weighted sum:
\begin{equation}
\mathbf{W} = \mathbf{W}_0 + \sum_{i \in \mathcal{P}} G(x)_i \cdot \mathbf{W}_i,
\end{equation}
where $\mathbf{W}_0$ represents the pre-trained frozen weights. Experts with smaller kernels focus on local, fine-grained patterns derived from the meta-knowledge, while experts with larger kernels capture global, long-range semantic dependencies. In our implementation, we typically set $k=2$, allowing the model to jointly leverage fine-grained local patterns (from small kernels) and global semantic dependencies (from large kernels).

\subsection{Adaptive LoRA Pruner}
In multi-task scenarios, facilitating the reuse of meta-knowledge across diverse tasks is critical for maximizing parameter utility. While standard MoE-LoRA variants assign independent LoRA modules as experts, this disjoint approach hinders the learning of universal representations. To overcome this, we propose the Adaptive LoRA Pruner, a mechanism that dynamically tailors the active parameters of the meta-knowledge base to match the granularity required by the current task scale. Specifically, for an expert $i$ requiring a target granularity $(h_t, w_t)$, we do not utilize the full meta-matrices. Instead, we slice the foundational matrices $\mathbf{B} \in \mathbb{R}^{H_{max} \times R}$ and $\mathbf{A} \in \mathbb{R}^{R \times W_{max}}$ into $\mathbf{B}_{:h_t, :}$ and $\mathbf{A}_{:, :w_t}$ respectively, where $R$ denotes the shared latent rank. This generates a scale-specific meta-seed:
\begin{equation}
\mathbf{Z}{meta}^{(t)} = \mathbf{B}{:h_t, :} \mathbf{A}{:, :w_t},
\end{equation}
consequently, the resulting $\mathbf{Z}_{meta}^{(t)}$ is a matrix of size $h_t \times w_t$, effectively capturing a sub-region of the meta-knowledge space calibrated to the target scale. Finally, $\mathbf{Z}_{meta}^{(t)}$ is projected by a scale-specific deconvolutional kernel $\mathcal{K}_i$ to produce the weight $\mathbf{W}_i$, whose dimensions are strictly consistent with the pre-trained weights of the $i$-th target layer.
% The parameterization of the $i$-th expert branch is then reformulated as:
% \begin{equation}
% \mathbf{W}_i = \operatorname{Deconv}(\mathbf{Z}_{meta}^{(t)}; \mathcal{K}_i),
% \end{equation}
% where $\mathbf{W}_i$ captures feature dependencies at a scale determined by the specific depth of the accessed meta-knowledge and the kernel size of $\mathcal{K}_i$. 

% However, this nested sharing mechanism introduces an update frequency imbalance. The parameters within the lower-indexed dimensions of the meta-knowledge subspace are active in every forward pass across all scales, leading to significantly more frequent gradient updates compared to the task-specific higher dimensions. To prevent this universal knowledge from being overwritten and to ensure stable convergence, We introduce a dimension-aware scaling factor similar to that of MoRE \cite{zhang2025moremixturelowrankexperts}. Let $d_t$ be the effective output dimension of expert $i$, and $T$ be the total number of tasks. The balanced forward process for the EPT layer is defined as:
% \begin{equation}
% \mathbf{L} = \mathbf{W}_0 \mathbf{x} + \sum_{i \in \mathcal{P}} G(x)_i \cdot \frac{d_t}{T} \cdot (\mathbf{W}_i \mathbf{x}),
% \end{equation}
% where $G(x)_i$ is the gating score. This mechanism acts as a regularizer that prioritizes the stability of universal patterns while allowing for larger update magnitudes for task-specific, high-dimensional features.
To ensure robust multi-task learning, we first address the data imbalance issue common in large-scale benchmarks. Following prior work \cite{zhang2025moremixturelowrankexperts}, we employ a Balanced Data Sampling strategy. For a set of $T$ tasks, we ensure that each task $t \in \{1, \dots, T\}$ is sampled with equal probability $P_t = 1/T$, regardless of its original dataset size $|D_t|$. While this strategy balances task-level contribution, it exacerbates the update frequency imbalance within our nested EPT layer. Under uniform task sampling, the parameters in the shared meta-knowledge subspace (lower-indexed dimensions) are updated in every training step (frequency $f_{shared} = 1$), whereas task-specific parameters (higher-indexed dimensions $d_t$) are only updated when their corresponding task is sampled (frequency $f_{specific} = 1/T$).To bridge this $T$-fold gap in update frequency and stabilize the optimization landscape, we introduce the dimension-aware scaling factor $d_t/T$. The forward pass is defined as:
\begin{equation}
\mathbf{L} = \mathbf{W}_0 \mathbf{x} + \sum_{i \in \mathcal{P}} G(x)_i \cdot \frac{d_t}{T} \cdot (\mathbf{W}_i \mathbf{x}),
\end{equation}
here, the factor $1/T$ serves as a frequency compensator: it scales down the per-step gradient magnitude of high-frequency shared dimensions to prevent them from being overwritten by rapid oscillations, while the $d_t$ term accounts for the increasing capacity of task-specific experts. This ensures that over the entire training trajectory, the accumulated gradient energy remains balanced across all dimensions of the meta-knowledge subspace, facilitating stable convergence in complex multi-task scenarios.

\begin{table*}
	\small
	% \scriptsize 
	\centering
    \renewcommand\arraystretch{1.2}
	\begin{tabular}{lcccccccccc}
		\toprule
		\textbf{Methods} & \textbf{params/task} & \textbf{MNLI} & \textbf{QQP} & \textbf{QNLI} & \textbf{SST-2} & \textbf{STS-B} & \textbf{MRPC} & \textbf{RTE} & \textbf{CoLA} & \textbf{AVG} \\
		\midrule
		Finetuning & 28M & 85.7 & 91.1 & 92.0 & 92.5 & 88.8 & 90.2 & 75.4 & 54.9 & 83.8 \\
		Adapters & 1.8M &86.3 & 90.5 & 93.2 & 93.0 & 89.9 & 90.2 & 70.3 & 61.5 & 84.4 \\
		PT & 9.6k & 85.6 & 90.6 & 93.2 & 93.9 & 89.9 & 86.3 & 67.6 & 55.3 & 82.8 \\
		$LoRA_{r=8}$ & 0.39M & 85.8 & 89.2 & 93.1 & 93.2 & 90.4 & 89.9 & 76.3 & 62.8 & 85.1 \\
		$LoRA_{r=16}$ & 0.78M & 84.9 & 89.6 & 93.0 & 93.7 & 90.4 & 88.7 & 80.6 & 63.9 & 85.6 \\
		\midrule
		HyperFomer & 638K & 85.7 & 90.0 & 93.0 & 94.0 & 89.7 & 87.2 & 75.4 & 63.7 & 84.8 \\
		MPT & 10.5K & 84.3 & 90.0 & 93.0 & 93.3 & \ccr{90.4} & 89.2 & \ccr{82.7} & 63.5 & 85.8 \\
		MultiLoRA & 1.56M & 85.9 & 89.7 & 92.8 & \ccr{94.5} & 89.8 & 88.2 & 80.6 & 66.9 & 86.0 \\
		MixLoRA & 1.49M & 85.8 & 90.0 & 92.9 & 93.7 & \cco{90.3} & 89.2 & 78.4 & 67.2 & 85.9 \\
		MOELoRA & 0.81M &\cco{86.3} & \ccr{90.4} & \cco{93.2} & \cco{94.2} & 89.8 & \ccr{90.7} & 79.9 & 65.3 & \cco{86.2}$_{\pm 0.15}$ \\
        MoRE & 0.81M & 85.6 & \cco{90.2} & 93.1 & 93.9 &89.9 & \ccr{90.7} & 77.7 & \cco{68.7} & \cco{86.2}$_{\pm 0.03}$ \\
		\midrule
        EPT & 0.41M & \ccr{86.4} & \cco{90.2} & \ccr{93.6} & \ccr{94.5} & 90.0 & \ccr{90.7} & \cco{82.0} & \ccr{68.9} & \ccr{87.0}$_{\pm 0.05}$\\
		% \midrule
		% LLaMA2-LoRA & 2.5M & 86.9 & \underline{88.6} & 93.5 & 96.2 & 90.2 & \textbf{92.6} & \underline{89.2} & 65.0 & 87.8 \\
		% LLaMA2-MultiLoRA & 10M & \underline{87.6} & 85.0 & 93.4 & \underline{96.7} & \underline{92.2} & 88.7 & 87.8 & \underline{72.4} & \underline{88.0} \\
		% LLaMA2-MixLoRA & 12.2M & 86.8 & 88.1 & \underline{93.6} & 96.0 & 91.3 & 88.2 & 87.1 & \textbf{73.2} & \underline{88.0} \\
		% LLaMA2-MOELoRA & 5M & 87.0 & 87.6 & 91.4 & 96.3 & \textbf{92.4} & \underline{91.2} & 87.8 & 64.4 & 87.3 \\
		% LLaMA2-MoRE & 5M & \textbf{89.4} & \textbf{89.0} & \textbf{94.4} & \textbf{96.9} & \underline{92.2} & 89.2 & \textbf{92.1} & 66.9 & \textbf{88.8} \\
		% \midrule
  %       LLaMA2-EPT \\
		\bottomrule
	\end{tabular}
	\caption{Performance on GLUE benchmark. For STS-B, we report Pearson correlation coefficients. For CoLA, we report Matthews correlation. For all other tasks, we report Accuracy. The \colorbox{red!25}{best score}, and \colorbox{orange!25}{second best score} are red, and orange, respectively. Subscripts denote the standard deviation of the average score over three runs.}
	\label{tab:overall_results}
\end{table*}
\subsection{Task Embedding}
Although existing MoE approaches have achieved notable progress in multi-task learning, they often fail to effectively disentangle the shared meta-knowledge from task-specific unique features. This limitation creates a bottleneck for the performance of PEFT in multi-task scenarios. To overcome this limitation, we propose introducing learnable task embeddings to explicitly model the latent correlations and discrepancies among tasks.

Specifically, we parameterize the set of all tasks as a matrix $\mathbf{E}=\{\mathbf{e}_1, \mathbf{e}_2, \dots, \mathbf{e}_T\}$, where $\mathbf{e}_i$ denotes the latent task prototype of the $i$-th task. 
To endow these embeddings with explicit semantic capabilities, we design a task-aware Contrastive Learning objective. Let $\mathbf{f}_i$ be the feature of sample $x_i$ and $\mathbf{e}_{t_i}$ be the corresponding task embedding. We define the temperature-scaled similarity score as $s_{i,k} = \operatorname{sim}(\mathbf{f}_i, \mathbf{e}_k) / \tau$. The contrastive loss is formulated as:
\begin{equation}
\mathcal{L}_{con} = - \frac{1}{M} \sum_{i=1}^{M} \log \frac{e^{s_{i, t_i}}}{\sum_{k=1}^{T} e^{s_{i, k}}},
\end{equation}
where $M$ is the batch size and $T$ is the total number of tasks. This objective essentially maximizes the mutual information between sample features and their task prototypes. Essentially, this loss function maximizes the mutual information between sample features and their corresponding task embeddings while pushing away irrelevant task embeddings, thereby compelling $\mathbf{E}$ to capture the intrinsic properties of the tasks. Furthermore, to ensure the generative capability of the model, we jointly optimize the standard generation task loss:
\begin{equation}
\mathcal{L}_{gen} = - \sum_{j=1}^{L} \log P(y_j | y_{<j}; \mathbf{x}, \mathbf{e}_t),
\end{equation}
where $y$ denotes the ground truth tokens. By jointly optimizing $\mathcal{L}_{total} = \mathcal{L}_{gen} + \lambda \mathcal{L}_{con}$, we achieve high-quality task representation learning in an unsupervised setting, significantly enhancing the robustness of multi-task fine-tuning.

\section{Experiment}
\subsection{Experimental Settings}
We use PyTorch \cite{10.5555/3454287.3455008} to implement all the algorithms. LLaMA2-7B \cite{touvron2023llama} and T5-base \cite{JMLR:v21:20-074} are employed as our foundational backbones. Optimization is performed using the AdamW optimizer with a peak learning rate of $3 \times 10^{-4}$, accompanied by a linear decay schedule and a 500-step warmup phase. All models are trained for 5 epochs with a global batch size of 32 and a maximum input sequence length of 128 tokens. For the EPT-specific hyperparameters, we set the contrastive loss weight $\lambda$ to 0.1 and the softmax temperature $\tau$ to 0.05. specific hyperparameters are detailed in Appendix \ref{Hyperparameters}. Computational resources consist of a single NVIDIA Tesla A100 GPU for T5-base and a cluster of three NVIDIA Tesla A800 GPUs for LLaMA2-7B.
\subsubsection{Datasets}
For evaluation, we adopt the GLUE benchmark \cite{wang-etal-2018-GLUE}, a widely recognized benchmark for natural language understanding, including CoLA \cite{warstadt-etal-2019-neural}, SST-2 \cite{socher-etal-2013-recursive}, MRPC \cite{dolan-brockett-2005-automatically}, QQP \cite{wang-etal-2018-GLUE}, STS-B \cite{wang-etal-2018-GLUE}, MNLI \cite{williams-etal-2018-broad}, QNLI \cite{rajpurkar-etal-2016-squad} and RTE \cite{10.1007/11736790_9}. Additionally, we included datasets like BoolQ \cite{clark-etal-2019-boolq}, OBQA \cite{mihaylov-etal-2018-suit}, and ARC \cite{clark-etal-2019-boolq} to test commonsense reasoning abilities. We present the dataset statistics in the Appendix~\mbox{\ref{A1} and \ref{A2}}.
\begin{table*}
	\small
	% \scriptsize 
	\centering
    \renewcommand\arraystretch{1.2}
	\begin{tabular*}{\textwidth}{@{\extracolsep{\fill}}l|ccccccc}
		\toprule
		\textbf{Methods} & \textbf{params/task} & \textbf{BoolQ}  & \textbf{OBQA} & \textbf{ARC-E} & \textbf{ARC-C}& \textbf{AVG} \\
		\midrule
        LoRA&2.1M& 74.0 &74.0& 80.9& 63.5 & 73.1\\
		MultiLoRA & 10M &\ccr{76.5} & 68.2 & \cco{81.2} & 61.9 & 72.0\\
		MOELoRA & 4.5M&73.3& 67.8&71.5&57.5&67.5 \\
    		MoRE & 4.5M & 74.7 & \ccr{80.5} & 80.0 & \cco{64.5} & \cco{74.9}\\
		\midrule
        EPT & 3.3M & \cco{76.1} &\cco{78.4}&\ccr{81.4}&\ccr{66.2}&\ccr{75.5}\\
		\bottomrule
	\end{tabular*}
	\caption{Accuracy of all methods on Commonsense Reasoning tasks. The backbone is Llama2-7B.}
	\label{tab:commonsense}
\end{table*}

\subsection{Overall Performance}
As shown in Table \ref{tab:overall_results} and Table \ref{tab:commonsense}, the performance on the GLUE benchmark and commonsense reasoning tasks indicates that the EPT method demonstrates exceptional results in few-parameter fine-tuning multi-task scenarios, outperforming the LoRA series of methods. In the GLUE task, we adopt T5-base as the backbone. Specifically, with only 0.41M parameters per task, EPT achieved an average score of 87.0\% across eight GLUE tasks. It attained the best performance on six of these tasks—MNLI, QNLI, SST-2, MRPC, RTE, and CoLA—significantly surpassing all other comparative methods, including those with higher parameter efficiency as well as advanced methods with comparable or greater parameter counts. This suggests that EPT achieves optimal overall performance while maintaining high parameter efficiency. The results show that by employing a task-specific embedding mechanism and a multi-scale pyramid projection mechanism, EPT can efficiently manage task information and accurately capture the low-dimensional latent structures across different tasks, thereby enhancing model performance without excessive parameter tuning. In contrast, parameter-efficient fine-tuning baseline methods, which lack mechanisms for integrating shared knowledge and require larger amounts of training data, perform poorly on small datasets. Although multi-task baseline methods consider knowledge sharing, they fail to adequately distinguish subtle differences among tasks, resulting in performance that lags behind EPT.

In commonsense reasoning tasks, we employ the larger LLAMA-2-7B as the backbone model. As the model parameters increase, EPT achieves the highest accuracy while using only 3.3M parameters per task, demonstrating its robustness across diverse scenarios. Compared to simple ensemble methods such as MixLoRA or MoELoRA, MoRE proves more effective at handling the complex and nuanced demands of commonsense reasoning tasks.

\begin{figure*}[t]
  \centering
  \begin{subfigure}{0.49\linewidth}
    \centering
    \includegraphics[width=\linewidth]{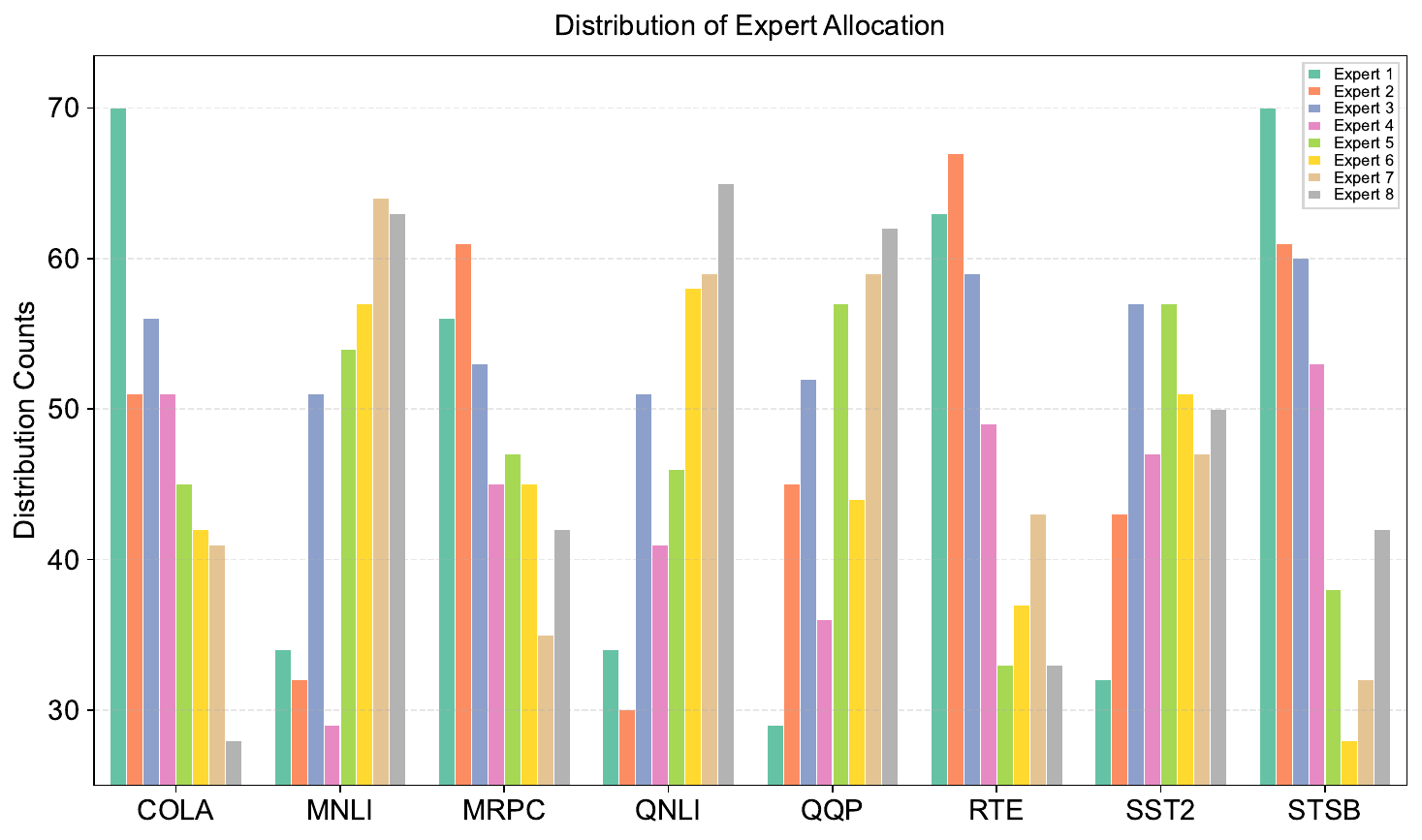}
    \caption{}
    \label{fig:accuracy}
  \end{subfigure}
  \hfill
  \begin{subfigure}{0.49\linewidth}
    \centering
    \includegraphics[width=\linewidth]{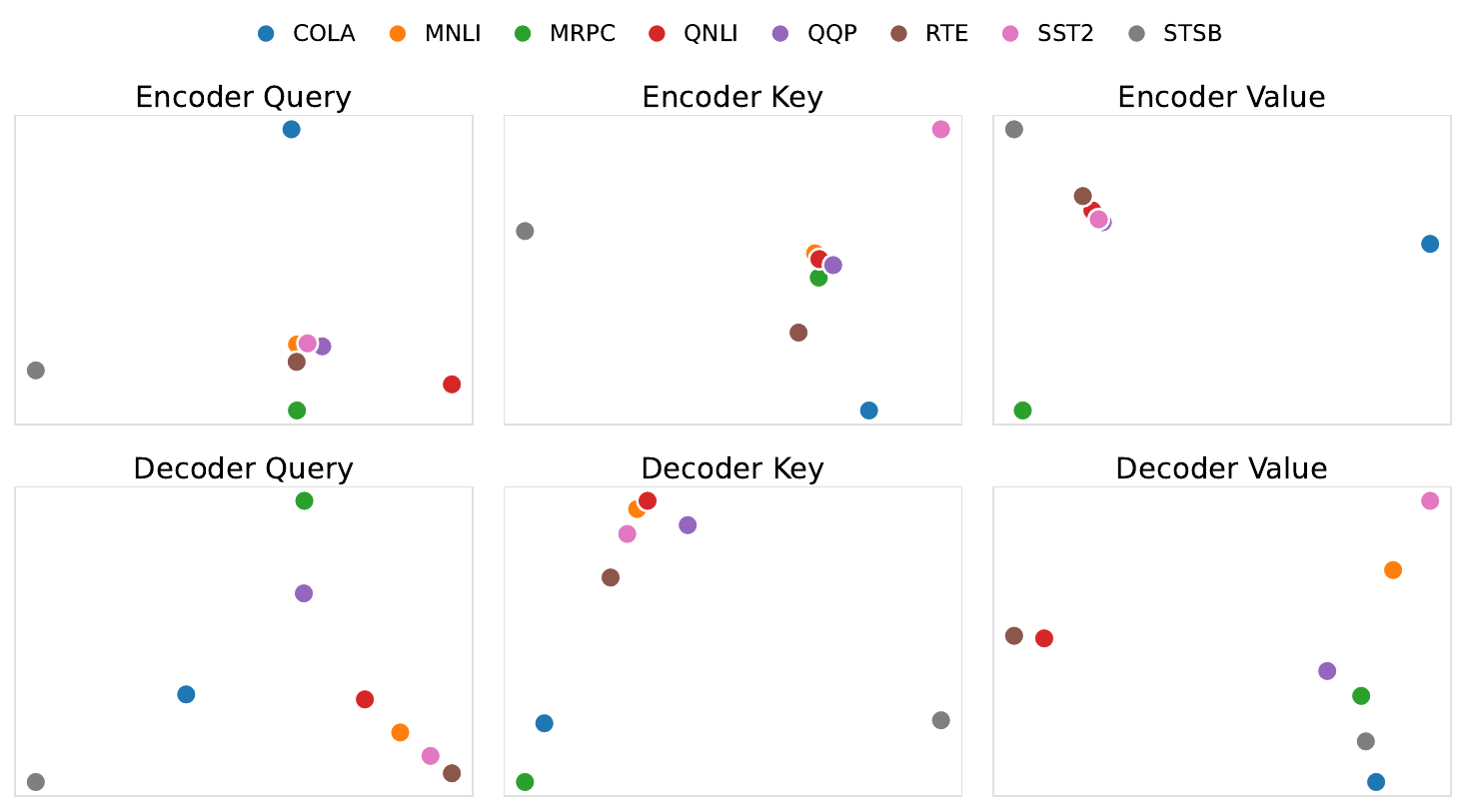}
    \caption{}
    \label{fig:max_layer}
  \end{subfigure}
  \caption{We present a visualization of the dataset analysis. Figure (a) shows the activation of each expert during the GLUE benchmark testing. Figure (b) displays the task embeddings from the final layer of the self-attention module.}
  \label{fig:combined}
\end{figure*}

\begin{table*}
    \small
    \centering
    \renewcommand\arraystretch{1.2}
    \begin{tabular*}{\textwidth}{@{\extracolsep{\fill}}l|ccccccccc }
        \toprule
        \textbf{Methods} & \textbf{MNLI} & \textbf{QQP} & \textbf{QNLI} & \textbf{SST-2} & \textbf{STS-B} & \textbf{MRPC} & \textbf{RTE} & \textbf{CoLA} & \textbf{AVG} \\
        \midrule
        \textbf{EPT-2}& 86.0 &\ccr{90.4} & 93.2 & \ccr{94.8} &90.2 & 90.2 & 83.5 & 63.7 & \cco{86.5} \\
        \textbf{EPT-4} &86.1&\ccr{90.4}&93.2&94.4&89.8&88.2&82.0&65.7&86.2\\
        \textbf{EPT-6} & 85.9&90.3&93.4&93.7&\cco{90.3}&\ccr{90.7}&74.8&\cco{68.0&}85.9\\
        \textbf{EPT-8} & \cco{86.2} & \ccr{90.4}& \cco{93.5} & 94.3&\ccr{90.4}&89.2&79.1&67.6&86.3\\
        \textbf{EPT-2468} &  \ccr{86.4} & 90.2 & \ccr{93.6} & \cco{94.5} & 90.0 & \ccr{90.7} & \ccr{82.0} & \ccr{68.9}& \ccr{87.0}\\
        \bottomrule
    \end{tabular*}
    \caption{Ablation Experiment Results. We conduct ablation experiments on initialization methods (AB init), Top-K, and the Adaptive LoRA Pruner (ALP) modules.}
    \label{tab:ept}
\end{table*}
\subsection{Expert Allocation Analysis}
After fine-tuning each task, we analyzed the distribution of expert weights across all layers. Experts 1-8 represent gradually increasing deconvolution dimensions, as shown in Figure 2(a). The analysis results indicate that different tasks typically correspond to different experts with the highest activation counts, which is largely consistent with our design philosophy. Furthermore, the figure reveals that the model learns to employ more advanced experts for larger, more challenging datasets. For instance, the large datasets QNLI and QQP predominantly utilize Expert 8, while MNLI relies more on Expert 7. In contrast, smaller and simpler datasets like STSB and RTE activate lower-level experts, such as Experts 1 and 2, more frequently. This confirms that EPT can effectively assign suitable experts to different tasks, thereby enhancing multi-task processing capabilities.
\subsection{Task Embeddings Analysis}
In this section, we explore the critical role of task embeddings in EPT's selection of experts. We visualize these embeddings using Principal Component Analysis (PCA) from the final layer of the self-attention module, as shown in Figure 2(b). The results reveal clear clustering patterns for similar tasks (such as QNLI and MNLI) and significant separation for tasks with notable differences (like STSB and CoLA). This is because QNLI and MNLI are both natural language inference tasks with similar formats and objectives, whereas STSB (a semantic textual similarity regression task) and CoLA (a linguistic acceptability judgment task) differ fundamentally in their required reasoning. These findings confirm that EPT can effectively distinguish and relate tasks through task embeddings, thereby optimizing the expert selection mechanism and enhancing the overall performance of EPT.
\begin{table*}
    \small
    \centering
    \renewcommand\arraystretch{1.2}
    \begin{tabular}{ccc|ccccccccc}
        \toprule
        \textbf{AB init}& \textbf{Top-K}& \textbf{ALP}& MNLI & QQP & QNLI & SST-2 & STS-B & MRPC & RTE & CoLA & AVG \\
        \midrule
        \ding{55}& \ding{55}& \ding{55} &85.5&90.0&92.7&93.7&89.0&90.2&81.3&65.7&86.0\\
        \ding{55}& \ding{55}& \ding{51} & 85.6 &\cco{90.2} & 93.1 & 93.9 &89.9 & \ccr{90.7} & 77.7 & 68.7 & 86.2 \\
        \ding{51} & \ding{55} & \ding{51} & \cco{86.3} & \ccr{90.4}& 93.2 & 94.3&90.2& 88.7& 80.6& 68.1& 86.5\\
        \ding{51} & \ding{51}  & \ding{55} & 86.1 & 90.0& 93.5 & 93.9&\ccr{90.4}&88.7&\ccr{82.0}&65.2&86.2\\
        \ding{55} & \ding{51}  & \ding{51} &86.2&90.2&93.4&\ccr{94.8}&\cco{90.3}&90.2&79.1&\ccr{69.4}&\cco{86.7}\\
        \ding{51} & \ding{51} & \ding{51} &  \ccr{86.4} & \cco{90.2} & \ccr{93.6} & \cco{94.5} & 90.0 & \ccr{90.7} & \ccr{82.0} & \cco{68.9}& \ccr{87.0}\\
        \bottomrule
    \end{tabular}
    \caption{Ablation Experiment Results. We conduct ablation experiments on initialization methods (AB init), Top-K, and the Adaptive LoRA Pruner (ALP) modules.}
    \label{tab:abl}
\end{table*}
\subsection{EPT Analysis}
A core motivation of EPT is that tasks of varying complexities require different representational granularities. To verify the necessity of the multi-scale pyramid structure, we conduct a comparative analysis by fixing the expert dimensions. Specifically, we compare our \text{EPT}-2468 configuration (with expert dimensions $\{2, 2, 4, 4, 6, 6, 8, 8\}$) against four baselines where all experts are assigned identical dimensions: \text{EPT}-2, \text{EPT}-4, \text{EPT}-6, and \text{EPT}-8. In all settings, the total number of experts is maintained at eight to ensure a fair comparison of architectural influence. As illustrated in our experimental results, the \text{EPT}-2468 configuration consistently outperforms all counterparts across the majority of multi-task benchmarks. We observe that while \text{EPT}-8 provides high capacity for linguistically complex tasks (e.g., CoLA), it often suffers from over-parameterization on simpler tasks (e.g., RTE), leading to suboptimal generalization. Conversely, \text{EPT}-2 maintains high efficiency but lacks the expressive power required for deep semantic reasoning. The superiority of EPT stems from its structural diversity. By projecting the shared meta-knowledge subspace into a multi-dimensional pyramid, EPT creates a more flexible hypothesis space. The router is not merely choosing between redundant experts of the same scale, but is dynamically allocating the most appropriate resolution of features for each task. This allows the model to capture fine-grained syntactic patterns via low-dimensional projections while simultaneously leveraging high-dimensional projections for global semantic abstraction. These results confirm that the EPT effectively mitigates negative transfer by providing a specialized capacity for diverse task demands, which a homogeneous expert pool fails to achieve.

\subsection{Ablation Study}
In this section, we conduct a comprehensive ablation study to systematically evaluate the contribution of each component in our framework. All experiments are performed under the same set of hyperparameters. The results are shown in Table \ref{tab:abl}, where the method that simultaneously adopts AB initialization, Top-K, and Adaptive LoRA Pruner achieves the best overall performance.

Effectiveness of AB init. Comparing the first and third rows, we observe that replacing the standard zero-initialization with our random Gaussian initialization for both $\mathbf{A}$ and $\mathbf{B}$ matrices (AB init) leads to a performance gain (from 86.2 to 86.5 in average score). This confirms our hypothesis that a non-degenerate latent representation at the onset of training provides a richer meta-knowledge seed, which is crucial for subsequent deconvolutional reconstruction.

Impact of Top-K Routing. The routing mechanism plays a pivotal role in multi-scale feature fusion. When comparing the third and sixth rows, the inclusion of Top-K routing (with $k=2$) consistently improves performance across most tasks, particularly on RTE (+1.4) and QNLI (+0.4). This suggests that adaptively selecting and combining experts with different kernel sizes allows the model to better balance local fine-grained patterns and global semantic dependencies.

Role of Adaptive LoRA Pruner. The ALP module is designed to maintain structural consistency and balance update frequencies. As shown in the fifth and sixth rows, the full EPT model (incorporating ALP) achieves the highest average score of 87.0\%. Notably, on the CoLA and SST-2 datasets, the presence of ALP significantly contributes to stability and accuracy. The results indicate that the dimension-aware scaling and the dynamic slicing mechanism in ALP effectively prevent universal meta-knowledge from being overwritten during task-specific adaptations.

\section{Conclusion}
In this paper, we introduced Expert Pyramid Tuning (EPT), a novel parameter-efficient fine-tuning framework that addresses the inherent limitations of uniform architectures in existing MoE-LoRA variants. By drawing inspiration from the multi-scale feature hierarchies in computer vision, EPT effectively captures the diverse granularities required by different linguistic tasks. Our approach decomposes task adaptation into a shared meta-knowledge subspace and a Pyramid Projection Mechanism, allowing the model to reconstruct features at varying scales through deconvolutional experts. To ensure structural consistency and training stability, we proposed the Adaptive LoRA Pruner (ALP), which utilizes a dimension-aware scaling factor to balance update frequencies across shared and task-specific parameters. Furthermore, our contrastive learning-based task embedding module significantly enhances expert routing by explicitly modeling the latent correlations and discrepancies among tasks. Extensive experiments on the GLUE benchmark and commonsense reasoning tasks demonstrate that EPT achieves SOTA performance, outperforming existing PEFT and MoE-LoRA baselines with high parameter efficiency. 
Notably, the re-parameterization capability of EPT achieves these performance gains with fewer training parameters. We believe that the concept of a parameter pyramid provides a promising direction for future research in multi-task adaptation and the deployment of Large Language Models in resource-constrained environments.
\section*{Limitations}
This study has two main limitations.
First, although the proposed expert pyramid introduces multi-dimensional projections within a single layer to capture diverse feature granularities, the specific configuration of these dimensions is currently treated as a static hyperparameter. Future research could explore dynamic dimension allocation or automated gating mechanisms to further refine the efficiency of this multi-dimensional projection.
Second, due to computational resource constraints, our evaluation is primarily focused on downstream fine-tuning tasks. While these results demonstrate the efficacy of the proposed method, its performance and stability during large-scale pre-training from scratch remain to be validated. Exploring the scalability of the expert pyramid in foundational model training is a crucial next step.

% \section*{Acknowledgments}

\bibliography{custom}
\clearpage
\appendix
\section{Datasets}
\subsection{NLU Datasets}
\label{A1}

For evaluation, we adaopt the GLUE benchmark \cite{wang-etal-2018-GLUE}, including CoLA \cite{warstadt-etal-2019-neural}, SST-2 \cite{socher-etal-2013-recursive}, MRPC \cite{dolan-brockett-2005-automatically}, QQP \cite{wang-etal-2018-GLUE}, STS-B \cite{wang-etal-2018-GLUE}, MNLI \cite{williams-etal-2018-broad}, QNLI \cite{rajpurkar-etal-2016-squad} and RTE \cite{10.1007/11736790_9}. We present the dataset statistics of GLUE in the following table \ref{tab:NLU}.

\begin{table}[!h]
    \centering
    \scalebox{0.77}{ 
    \renewcommand\arraystretch{1.5}
    \begin{tabular}{l|ccccc} 
    \toprule
      \textbf{Dataset} &  \textbf{Metric}  &  \textbf{\#Train} & \textbf{\#Valid} & \textbf{\#Test}& \textbf{\#Label}  \\ \midrule
        CoLA & Mcc&8.5k &1,043 &1,063 &2  \\
        SST-2& Acc &67k &872 &1.8k &2 \\
        MRPC & Acc&3.7k &408 &1.7k &2\\
        QQP  & Acc/F1&364k &40.4k &391k &2 \\
        STS-B& Corr&5.7k &1.5k &1.4k &1 \\
        MNLI & Acc(m/mm)&393k &20k &20k&3 \\
        QNLI & Acc&105k &5.5k &5.5k &2 \\
        RTE  & Acc&2.5k &277 &3k &2\\
    \bottomrule
    \end{tabular}}
    \caption{ Dataset Sizes and Evaluation Metrics in the GLUE Benchmark. "Mcc," "Acc," "F1," and "Corr" denote the Matthews correlation coefficient, accuracy, F1 score, and Pearson correlation coefficient, respectively. "Acc(m/mm)" indicates accuracy results for matched and mismatched datasets within MNLI.}
    \label{tab:NLU}
\end{table}

\subsection{Commonsense Datasets}
\label{A2}
We present the dataset statistics of commonsense reasoning in the table~\mbox{\ref{tab:com}}. Additionally, we included datasets like BoolQ \cite{clark-etal-2019-boolq}, OBQA \cite{mihaylov-etal-2018-suit}, and ARC \cite{clark-etal-2019-boolq} to test commonsense reasoning abilities.
\begin{table}[!h]
\centering
\scalebox{0.77}{
\renewcommand\arraystretch{1.5}
\begin{tabular}{l|ccccc}
\toprule
\textbf{Dataset} & \textbf{Metric} & \textbf{\#Train} & \textbf{\#Valid} & \textbf{\#Test}& \textbf{\#Label} \\
\midrule
BoolQ & Acc &9.4k &3.3k &3.2k &2 \\
OBQA & Acc &4.9k &500 &500 &4 \\
ARC-E & Acc &2.2k &570 &2.4k &4 \\
ARC-C & Acc &1.1k &299 &1.2k &4 \\
\bottomrule
\end{tabular}}
\caption{Dataset Sizes and Evaluation Metrics in the commonsense question-answering datasets, where BoolQ is a binary classification (yes/no) task, while OBQA and ARC are multiple-choice questions. ARC is divided into two subsets: Easy (ARC-E) and Challenge (ARC-C).}
\label{tab:com}
\end{table}

\section{Algorithm}
In this section, we provide a comprehensive breakdown of the Expert Parameter Pyramid (EPT) fine-tuning pipeline. The EPT framework is designed to bridge the gap between low-rank adaptation and multi-scale expert modeling by generating a hierarchical set of adapters from a shared meta-parameter space.

Algorithm \ref{alg:ept_flow} outlines the end-to-end training procedure for EPT. Unlike traditional LoRA \cite{hu2022lora} which applies a single fixed-rank update, EPT constructs a pyramid of expert weights $\mathbf{W}_i$ with varying dimensionalities. This allows the model to capture both coarse-grained task features and fine-grained linguistic nuances simultaneously.

\begin{algorithm*}[ht]
\caption{EPT Fine-tuning Pipeline}
\label{alg:ept_flow}
\begin{algorithmic}[1]
\renewcommand{\algorithmicrequire}{\textbf{Input:}}
\renewcommand{\algorithmicensure}{\textbf{Output:}}

\REQUIRE Pre-trained Model $f_{\phi}$ (frozen); Dataset $\mathcal{D}$; Initial EPT parameters $\Theta = \{A, B, \mathcal{K}, E, W_r\}$.
\ENSURE Optimized EPT parameters $\Theta^*$.

\STATE \textbf{Initialize:} $A, B \sim \text{Gaussian}$, $\mathcal{K} \leftarrow 0$, $E \leftarrow \text{Random}$
\STATE Freeze backbone parameters $\phi$

\WHILE{training}
    \FOR{each batch $\{(x, t, y)\} \subset \mathcal{D}$}
        \STATE // \textit{1. Parameter Pyramid Generation (Offline per batch)}
        \STATE $\mathbf{Z}_{meta} \leftarrow B \cdot A$
        \FOR{$i=1$ \TO $N$}
            \STATE $\mathbf{W}_i \leftarrow \text{Deconv}(\text{Slice}(\mathbf{Z}_{meta}, d_i); \mathcal{K}_i)$ \COMMENT{Build Expert Pyramid}
        \ENDFOR
        
        \STATE // \textit{2. Multi-task Forward Propagation}
        \STATE $\mathbf{h} \leftarrow f_{\phi}(x)$ \COMMENT{Backbone feature extraction}
        \STATE $\mathcal{L}_{con} \leftarrow \text{Contrastive}(\mathbf{h}, E, t)$ \COMMENT{Task embedding alignment}
        
        \STATE // \textit{3. Layer-wise EPT Adaptation}
        \STATE $\mathcal{P} \leftarrow \text{Top-k}(\text{Router}(x; W_r))$ \COMMENT{Identify task-appropriate scales}
        \STATE $\Delta \mathbf{h} \leftarrow \sum_{i \in \mathcal{P}} G(x)_i \cdot (\frac{d_i}{T}) \cdot (\mathbf{W}_i \mathbf{h})$ \COMMENT{Pyramid aggregation}
        
        \STATE // \textit{4. Prediction and Joint Optimization}
        \STATE $\hat{y} \leftarrow \text{Head}(\mathbf{h} + \Delta \mathbf{h})$ \COMMENT{Residual-style adaptation}
        \STATE $\mathcal{L}_{gen} \leftarrow \text{CrossEntropy}(\hat{y}, y)$
        \STATE $\mathcal{L}_{total} \leftarrow \mathcal{L}_{gen} + \lambda \mathcal{L}_{con}$
        
        \STATE // \textit{5. Parameter Update}
        \STATE $\Theta \leftarrow \Theta - \eta \nabla_{\Theta} \mathcal{L}_{total}$ \COMMENT{Update only EPT modules}
    \ENDFOR
\ENDWHILE
\RETURN $\Theta$
\end{algorithmic}
\end{algorithm*}

\section{Parameter Efficiency Analysis}
\label{Parameter}
In this section, we provide a quantitative analysis of the parameter efficiency of EPT compared to existing Mixture-of-Experts (MoE) based LoRA variants. For a fair comparison, we use the T5-base model as the backbone ($d=768$) and evaluate the additional trainable parameters per layer (excluding router parameters). We assume a standard configuration with $N=8$ experts and a hidden rank $r=8$.

Comparison with BaselinesTraditional MoE-LoRA architectures assign independent low-rank matrices to each expert. The total parameters per layer are calculated as $N \times (d \times r + r \times d)$. For $N=8$, this results in:
\begin{equation}
P_{\text{MoE-LoRA}} = 8 \times (768 \times 8 + 8 \times 768) = 98,304
\end{equation}
Recent variants like MoRE attempt to reduce redundancy by sharing parameters across experts. While this significantly lowers the overhead, it effectively reduces to a single set of shared low-rank matrices:
\begin{equation}
P_{\text{EPT}} = 768 \times 8 + 8 \times 768 = 12,288
\end{equation}
EPT EfficiencyIn contrast, EPT optimizes efficiency by decomposing the adaptation into a shared meta-knowledge subspace and a lightweight pyramid projection mechanism. Our parameter count consists of two parts:Shared Meta-knowledge Subspace: We employ a reduced-dimension subspace ($d_{sub}=384, r=8$), yielding $2 \times (384 \times 8) = 6,144$ parameters.Pyramid Projection Kernels: Instead of full matrices, EPT utilizes small deconvolutional kernels. For a pyramid with scales $s \in \{2, 4, 6, 8\}$, the parameters are $\sum 2 \times s_i^2 = 2 \times (2^2 + 4^2 + 6^2 + 8^2) = 240$. The total parameters for EPT are:
\begin{equation}
P_{\text{EPT}} = 6,144 + 240 = 6,384
\end{equation}

\section{Hyperparameters}
\label{Hyperparameters}
We employed the AdamW optimizer with a learning rate of $3 \times 10^{-4}$, a linear learning rate scheduler, and a warmup phase of 500 steps. The training was performed for 5 epochs with a per-device batch size of 8, resulting in a total effective batch size of 32. For parameter-efficient fine-tuning, we integrated LoRA adapters with a rank ($R$) of 8 and an alpha ($\alpha$) of 32, targeting all primary projection layers including $Q, K, V, O$ and MLP matrices. Furthermore, we implemented a Mixture-of-Experts (MoE) strategy using a Top-2 routing mechanism and a multi-scale expert kernel configuration of $[2, 2, 4, 4, 6, 6, 8, 8]$ to capture features across varying dimensions. The final model was selected based on the best performance achieved on the evaluation set during the training process.

\begin{table}[h]
\centering
\begin{tabular}{lc}
\toprule
\textbf{Hyperparameter} & \textbf{Value} \\ 
\midrule
Optimizer & AdamW \\
Learning Rate & $3 \times 10^{-4}$ \\
Weight Decay & 0.01 \\
Total Batch Size & 32 \\
Training Epochs & 5 \\
Max Sequence Length & 128 \\
Warmup Steps & 500 \\
LoRA Rank ($R$) & 8 \\
LoRA Alpha ($\alpha$) & 32 \\
Target Modules & q k v o up down \\
MoE Top-$k$ & 2 \\
Expert Kernel Sizes & [2, 2, 4, 4, 6, 6, 8, 8] \\
\bottomrule
\end{tabular}
\caption{Hyperparameter settings.}
\label{tab:hyperparams}
\end{table}

\end{document}